# Deep mineralogical segmentation of thin section images based on QEMSCAN maps

Jean Pablo Vieira de Mello[a], Matheus Augusto Alves Cuglieri[b], Leandro P. de Figueiredo[a], Fernando Bordignon[a], Marcelo Ramalho Alburquerque[b], Rodrigo Surmas[b] and Bruno Cavalcanti de Paula[b]

[a]*LTrace Geosciences*
[b]*Petrobras*

ABSTRACT

Interpreting the mineralogical aspects of rock thin sections is an important task for oil and gas reservoirs evaluation. However, human analysis tend to be subjective and laborious. Technologies like QEMSCAN® are designed to automate the mineralogical mapping process, but also suffer from limitations like high monetary costs and time-consuming analysis. This work proposes a Convolutional Neural Network model for automatic mineralogical segmentation of thin section images of carbonate rocks. The model is able to mimic the QEMSCAN mapping itself in a low-cost, generalized and efficient manner. For this, the U-Net semantic segmentation architecture is trained on plane and cross polarized thin section images using the corresponding QEMSCAN maps as target, which is an approach not widely explored. The model was instructed to differentiate occurrences of Calcite, Dolomite, Mg-Clay Minerals, Quartz, Pores and the remaining mineral phases as an unique class named "Others", while it was validated on rock facies both seen and unseen during training, in order to address its generalization capability. Since the images and maps are provided in different resolutions, image registration was applied to align then spatially. The study reveals that the quality of the segmentation is very much dependent on these resolution differences and on the variety of learnable rock textures. However, it shows promising results, especially with regard to the proper delineation of minerals boundaries on solid textures and precise estimation of the minerals distributions, describing a nearly linear relationship between expected and predicted distributions, with coefficient of determination ($R^2$) superior to 0.97 for seen facies and 0.88 for unseen.

## CRediT authorship contribution statement

**Jean Pablo Vieira de Mello:** Coding, writing, image registration . **Matheus Augusto Alves Cuglieri:** Methodology designing, data preparation, geological and sedimentological consulting. **Leandro P. de Figueiredo:** Advisor, Methodology designing,image registration, article revision. **Fernando Bordignon:** Project management, study designing, Co-advisor. **Marcelo Ramalho Alburquerque:** Project management, Co-advisor. **Rodrigo Surmas:** Project management, study designing. **Bruno Cavalcanti de Paula:** Data preparation, Geological and sedimentological consulting.

## 1. Introduction

Understanding the mineralogy of rocks is an important geological and petrophysical task in the oil and gas industry, as it allows evaluating the quality and genesis of reservoir rocks (Saxena et al., 2021; Fu et al., 2023; Nath et al., 2023; Manzoor et al., 2023). A common preliminary step is obtaining a small rock sample and extracting from it a 30 μm thick thin section, which is visible via a light polarizer microscope (exhibiting different mineral properties based on the polarization angle) and a digital camera for registering and storing high-resolution RGB photographs/photomosaics

ORCID(s): 0000-0000-0000-0000 (J.P.V.d. Mello)

of the thin sections (Tang et al., 2020; Saxena et al., 2021; Dabek et al., 2023; Latif et al., 2022). However, human evaluation of thin sections is very time-consuming and prone to fatigue and subjectivity, demanding reliable automated analysis approaches as the amount of available data increases (Asmussen et al., 2015; Tang et al., 2020; Saxena et al., 2021; Latif et al., 2022; Yu et al., 2023; Koh et al., 2024).

Over time, new technologies emerged in order to tackle this issue. One of the most popular is the *Quantitative Evaluation of Minerals by Scanning Electron Microscopy* (QEMSCAN®), which enables the generation of mineralogical maps of rock sample surfaces by integrating two base techniques: Scanning Electron Microscopy (SEM) and Energy-dispersive Spectometry (EDS) (Ayling et al., 2012; Ali et al., 2023). SEM consists in exciting the surface with an electron beam and mapping the response signals along the scanned region, which provides qualitative insights about the surfaces topography, chemical composition, crystalline structure and electrical properties. In turn, EDS quantifies the intensities of the X-rays emitted by the excited regions, which are characteristically different for different minerals. Then, the backscattered electron and X-ray signal levels are compared to the predefined in a mineral database, identifying the mineral in each region. Since the QEMSCAN methodology relies on the physical and chemical properties of the rocks, it softens the human analysis weaknesses. However, it also suffers from important limitations, such as high costs, complex operation and a still high time-consumption (Koh et al., 2021; Latif et al., 2022).

With the recent advances in Machine Learning (ML) and Computer Vision (CV), image-based approaches have been suggested in order to build simple, cheap and efficient mineral mapping for thin section images. Izadi et al. (2015) proposed an algorithm to cluster similar pixels into the same mineral category based on their RGB and HSV values, which was further explored by Izadi et al. (2020). Naseri and Rezaei Nasab (2023) also use RGB and HSV information as features, but they include texture characteristics as well and use the whole feature set as input to a SVM classifier (Boser et al., 1992; Cortes and Vapnik, 1995).

With the popularization of Deep Learning (DL) methods and, more specifically, Convolutional Neural Networks (CNN), the interest on DL solutions over traditional methods inspired several recent works. For instance, Saxena et al. (2021) and Nath et al. (2023) evaluated the semantic segmentation performance of well-known CNN architectures such as ResNet-18 (He et al., 2016), DeepLab V3+ (Chen et al., 2018) and U-Net (Ronneberger et al., 2015) on manually labeled thin section images. Koh et al. (2021), in the other hand, adopted an instance segmentation approach in order to isolate grains mainly composed of a specific mineral, building upon the consolidated networks Mask R-CNN (He et al., 2017) and SOLOv2 (Wang et al., 2020). These proposed methods tend be to be promising, but share a common issue: the dependence on manual data labeling, which is a very costly task.

To tackle this, Tang et al. (2020) label only a fraction of the grains in each image (attributing a single mineral class for the entire grain) and "eliminate" the remaining ones by replacing them by synthetic pore space. Then, the U-Net architecture is trained based on this modified labeled dataset. In turn, Yu et al. (2023) suggest an image labeling methodology that consists of pre-segmenting the images using a superpixel algorithm and then manually labeling the superpixels, leveraging the automatic decisions on the grains boundaries and reducing the efforts on labeling every pixel of the image.

Despite such solutions for reducing efforts in manual labeling, it is still a requirement for such methods to succeed. Therefore, the use of QEMSCAN mineral maps as labels to the corresponding thin section images appears to hold significant potential, since they are generated automatically. However, it is still uncommon to find in the literature works which make use of this method. Very recently, a work presented by Vellappally et al. (2024) adopted such approach for classifying thin section grains and showed promising results. In this work, we propose a method to combine the quality of the QEMSCAN mineralogical analysis with the advantages of image segmentation for a wide range of thin section textures, whether dense or granular. A U-Net-based network is trained to mimic the results of QEMSCAN applied to thin section images, enabling fast, low-cost and non-subjective mineralogical segmentation. Significant resolution differences between thin section images and QEMSCAN maps are handled through image registration using the open-source software GeoSlicer.

## 2. Dataset

This work makes use of a dataset with 102 carbonate thin section high resolution images from 22 different wells from the Brazilian Santos Basin, consisting of micrometer-scale photographs taken under plane (PP) and cross (XP) polarized lights. Each thin section's corresponding QEMSCAN map was used as the target image to be approximated by the network. More details about QEMSCAN maps are discussed in the next section.

In order to enable the training of a deep segmentation model, some preprocessing steps were applied to the available data, including *image registration* (Section 2.2) for spatial alignment, *segment of interest (SOI) delimitation* (Section 2.3) for information filtering and *chunking* (Section 2.4) for increasing data diversity. Figure 1 shows an overview of the data aspect and preprocessing.

### 2.1. QEMSCAN maps

Each different color shown in the maps refers to a different element from the thin section, including pores and different types of minerals. Although pores comprehend the absence of minerals, this work will treat each of these elements as different *mineral phases*, for simplicity.

The predominance of cyan and yellow colors in Figure 1 example refers to high levels of, respectively, calcite and quartz in the sample. Thin portions of blue (dolomite) and black (pores) can be also seen. However, lots of other mineral phases are represented in small and sparse portions, like the carbon rich material represented as tiny magenta islands in Figure 2. In fact, the example map characterizes 13 different mineral phases. The white border describes the class "Others", also designated to some occurrences among the minerals that are not attributed to a specific type of mineral phase.

Acquiring the QEMSCAN maps is a time-consuming task, taking several hours for a single sample. The balance between efficiency and quality in the mapping process relies on the *SEM scanning resolution*, i.e., the set distance interval between sampling points in a scanning mesh. The maps used in this work were acquired under a sampling resolution of 10 µm, which means that each pixel of a map describes the mineral phase detected in a distance of 10 µm from the adjacent ones in both *x* and *y* directions in a Cartesian plan. This results in a map which do not label every single pixel of the corresponding thin section images, but only a fraction of them in regular intervals. This relative low pixel resolution justifies the aspect of the map shown in Figure 2.

## 2.2. Image registration

As illustrated in Figure 1, the images were not necessarily provided in the same orientation and, as stated in Section 2.1, the QEMSCAN map is generated in resolution that differs from the thin section's. For this reason, each thin section had to be subjected to *image registration*, i.e., the spatial alignment of all three images. Basically, the process consists of applying geometric transformations such as translation, rotation, scaling and/or deformation to the *moving image*, so that it becomes aligned to a *fixed image* based on similar features (Oliveira and Tavares, 2014).

All provided thin sections had their PP and XP images already registered with each other or had their orientation differing by a simple 90°-multiple rotation, which just needed to be undone for any of them. Applying such transformation on the QEMSCAN map could also make them have similar orientations. In the Figure 1 example, the most appropriate alternatives would be either rotating PP 90° clockwise or both XP and QEMSCAN map counterclockwise. However, this is not enough to register the QEMSCAN map to the other images. In fact, the maps registration demanded a manual transformation process, which was performed using the platform for digital rocks processing GeoSlicer[1], built upon the open-source software 3D Slicer[2].

First of all, the images should be loaded into GeoSlicer environment and scaled according to the real dimensions they represent. In the case of the PP and XP images, this scale is approximately 1.32 µm/pixel. For the QEMSCAN maps, however, the default value (from the SEM sampling resolution) is 10 µm/pixel. This results in thin section images and QEMSCAN maps with different pixel resolutions but the same real-world size. For instance, consider a thin section having PP and XP images with side of 18,500 pixels and its corresponding map having a 2,500 pixels side. In millimeters, both thin section and map would share the same side size:

- Thin section: 18,500pixels x 1.32um/pixel = 25mm
- QEMSCAN map: 2,500pixels x 10um/pixel = 25mm

[1] https://github.com/petrobras/GeoSlicer
[2] https://www.slicer.org/

Once the images are loaded and properly scaled, either of PP or XP image could be subjected to the registration process as the fixed image and the QEMSCAN map should enter as the moving image. Although the scaled images' dimensions become close to each other's, they usually are still notably unregistered at this point. Figure 3 exemplifies a scenario with the images loaded and scaled but not properly registered yet.

The next step is to manually set a pair of landmarks so that one lies in an arbitrary location of the thin section image and its peer lies as near as possible to the corresponding location in the QEMSCAN map. Figure 4 illustrates an appropriate landmark placement.

Then, this process is repeated until the registration quality becomes acceptable. For each pair added, the software internally calculates the proper transformation matrices to be applied to the moving image so that its marked locations match those from the fixed image. Overall, at least four pairs of landmarks, as far as possible from each other, were set for each thin section. However, ome samples (such as those highly granular)demanded additional landmarks, either far or close to the main ones. Generally, the registration tend to be better as more landmarks are added, but the manual effort increases too. Figure 5 shows the result of the registration process performed with four landmark pairs. Notice how the registered map looks rotated compared to the original.

Although PP and QEMSCAN images seem to overlap perfectly in the figure, the intrinsic difficulty of manual registration combined with the significant differences between their original resolutions leads to a certain degree of mismatch, especially at the boundaries between mineral phases and, consequently, in highly granular regions/samples. For simplicity, this will be further referred as *the registration problem*. Figure 6 shows an example of such inaccuracies in the registration process.

### 2.3. Segment of interest (SOI)

A natural decision for the training process would be to ignore all the area from the thin section images covered by the QEMSCAN maps' borders or beyond them. However, some aspects of the dataset make this not viable:

- Simply discarding the maps' borders by value would also discard eventual occurrences of "Others" among other mineral phases;
- Some maps provide mineralogical information of thin section regions not exhibited in the PP and XP images;
- Some thin section images contain defective regions with stains or blurring.

To tackle these issues, the GeoSlicer platform was also used to delimit a *segment of interest* (SOI) for each thin section as a mask for the region with useful information. All the area from the images and QEMSCAN map not covered by SOI is discarded.

**Table 1**
Summary of the validation methods applied: *same* validates on non-trained portions of trained thin sections, while *split* validates on a different set of images than the used for training.

| Validation method | # training epochs | Train set | | Validation set | | Reason |
|---|---|---|---|---|---|---|
| | | Description | # chunks | Description | # chunks | |
| *same* | 300 | Largest portion of each thin section | 15,561 | Smallest portion of each thin section | 3,843 | Greater variability in training data |
| *split* | 75 | Largest thin section subset | 14,927 | Smallest thin section subset | 4,477 | Better notion of generalization for unknown thin sections |

## 2.4. Chunking

For obtaining greater data quantity, variety and processing ease, the images were divided into chunks with dimension of $1,000 \times 1,000$ pixels. First, the area outside the SOI's bounding box was discarded. Then, the chunking was performed on the largest possible center crop in which the chunks would fit entirely, preserving only chunks with at least 70% of area covered by SOI. This resulted in a dataset of 19,404 chunks. Figure 7 illustrates the process on the SOI superimposed on the images and map.

# 3. QEMSCAN-based mineralogical segmentation

The resulting chunks dataset comprehend a total of 35 mineral phases. Most, however, occur with a negligible frequency. For this reason, the aforementioned class "Others" was expanded to encompass all but the five most frequent phases. Then, the six final classes considered were: *Calcite (0 to 1% MgO)*, *Dolomite*, *Mg-Clay Minerals*, *Quartz*, *Pores* and *Others*. Figure 8 shows the original and adapted distributions of the mineral phases in the dataset.

The proposed segmenter is based on the Residual U-Net architecture for semantic segmentation (Kerfoot et al., 2019), which consists in the classical U-Net (Ronneberger et al., 2015) benefiting from some residual connections between layers (He et al., 2016). It encodes the input images into a deep low-resolution feature space and decodes the result to an output activation map with the same resolution as the inputs. In this work, the inputs are six-channel images stacking the RGB channels from PP and XP, the encoding-decoding process is conducted by strided convolutions and the outputs are maps with seven channels: one for each class plus one for the non-SOI area, so that a pixel's estimated class is determined by the channel with the strongest activation in its position (excluding non-SOI). Once the segmenter is trained with small augmented chunks, it is expected to be able to infer the mineralogy of entire thin sections. An overview of the network and method can be seen in Figure 9.

## 3.1. Training and validation

Given the relatively low number of available thin sections, two validation methods were tested, as shown in Table 1.

For the validation method *split*, 22 thin sections were picked for the validation set, with no expressive divergence in mineral phase distribution from the training set. In turn, the validation set for the method *same* considered the last 20% of the chunks from each thin section, generated in column-major order, constituting an intentionally biased experimentation.

The Dice coefficient (Milletari et al., 2016) was used as both loss and validation metric. It scores the overlap between the predicted and the target segmentations for each class in a one-*versus*-all basis, fitting well for cases with such high class imbalance. The eventual regions not covered by SOI were not considered for the calculations.

The models were trained on a Tesla V100 GPU with 32GB memory, preserving the model with the highest Dice score. The training was conducted considering a batch size of 16, a validation interval of 15 epochs and a different number of training epochs for each method, as shown in Table 1, for reasons discussed in Section 4. During training, the following augmentation processes were applied to each chunk for further data diversity:

- Random cropping of a $512 \times 512$ pixels area;
- Random horizontal and vertical flipping;
- Random rotation by a multiple of 90°.

In turn, the inference process for validation followed a sliding window fashion, considering a window size of $512 \times 512$ pixels and an overlap of 25% between windows, aggregating the prediction results by equally-weighted average. This procedure does not depend on the input size and preserve the training input resolution, so it can be applied not only for the validation chunks but also for entire thin sections. Figure 9b exemplifies a training and an inference iteration.

### 3.2. Evaluation

The trained model was evaluated not only based on the quality of segmentation at pixel level, but also on the similarity between the original mineral phases distribution and the one output by the segmenter. The former case was evaluated by the aforementioned Dice coefficient and the classes' confusion matrix, and the latter by the coefficient of determination $R^2$ and the Root Mean Squared Error (RMSE) between the original and predicted distributions.

## 4. Results and discussion

Figure 10 shows the behavior of the loss function (left) and Dice metric (right) for the models validated by the method *same*, in the upper row, and *split*, in the lower row. According to the loss curves, in both cases it is noticeable that both training and validation errors stabilize close to 0.7, which indicates that no significant overfitting or under-fitting affected the learning process. However, the *split* method presents two peculiar differences: convergence is way faster and, if the training proceeds for much longer, the training loss suffers an abrupt fall, which also decreases the validation loss but slightly decreases the Dice score too. In general, the Dice scores indicate somewhat limited levels of overlap between the reference (QEMSCAN map) and the predictions.

Figure 11 plots the linear correlations between the groundtruth and predicted distributions of each class in each validation thin section, as well as reveals the overall $R^2$ and RMSE. The greater the correlation is, $R^2$ is closer to 1, RMSE is closer to 0 and the lines slopes are closer to the unitary (45°). Despite the overlapping issues pointed by the Dice scores, the correlation graphics (Figure 11) show that the overall predicted distribution of the mineral phases is satisfactorily realistic, especially with regard to the validation method *same*. For the method *split*, the metric is slightly less pronounced but still numerically acceptable. Considering the classes individually, both methods seem to perform well on most, but present questionable performance on predicting the "Others" class. Also, predicting "Mg-Clay Minerals" challenges the *split* method. Similar behaviour is evident in the confusion matrices shown in Figure 12. Besides, there seems to be some concentration of predictions in some classes which appear frequently in the dataset, "Calcite (0 to 1%MgO)", "Dolomite" and "Quartz".

Finally, Figure 13 enables some visual assessment of the prediction quality of the models. The visual predictions also show some tendencies in regard to different aspects: the boundaries between mineral phases are smoother than the corresponding ones from the QEMSCAN map. Also, an overview of the images suggests that the methods yield predictions that are similar to the real map and to each other, while a closer look reveals many punctual divergences. In general, they tend to agree on regions with more regular textures and disagree on regions with higher variance of mineral phases occurrences.

Overall, the results presented indicate a performance that is promising but passive of improvements. The limitations detected can be attributed mainly to two important challenges: the registration problem and the low data variability. Due to the registration problem, the registered QEMSCAN maps provide wrong labeling to lots of pixels from their corresponding images, but their classes distributions are not affected by their spatial displacement. This means that if every single pixel of the thin section images was classified correctly, the QEMSCAN and the predicted maps would mismatch significantly, yielding low Dice scores and inaccurate confusion matrices although they were expected to be perfect, while the correspondence between the actual and predicted distributions would tend to remain very close to each other. This might indicate that the mineral phases patterns were well learnt by the models despite of the low Dice scores, which is further evidenced by the smoother and, therefore, more realistic, predicted boundaries. The problem can also explain the under-prediction of "Others" occurrences while over-predicting the group "Calcite (0 to 1%MgO)", "Dolomite" and "Quartz": since the pixels of the boundaries between mineral phases tend to be mislabeled, many of them are attributed to neighbor classes, affecting rare and predominantly-granular phases like "Others" and favoring more frequent ones.

Another issue to be considered refers to the fact that the number and variety of thin section images used for training are very small, limiting the models' generalization capacity. The method *same* demanded much more epochs to reach convergence as more rock textures needed to be recognized and, in return, the overall results showed to be superior, evidencing that the small set of thin sections excluded from the *split* training would be really impactful for the learning process. In fact, the behaviour of the loss curves of both methods seems to be the same except for the convergence time, and also the sudden drop might also happen with further *same* training (probably as a response from the learning rate scheduling), though not really improving the results. Also, the total distribution of a mineral phase distribution might be concentrated in few thin sections, limiting the learnable textures in which it occurs and yielding undesirable results. This seems to be case of "Mg-Clay Minerals" for the *split* method.

## 5. Conclusions

Both human and device-automated mineralogical analysis of rock thin sections are laborious and expensive tasks, demanding new practical solutions. By leveraging the power of consolidated Convolutional Neural Network architectures, this work presented the possibility of building an intelligent model for mineralogical segmentation of thin section images, trained to mimic the output of a QEMSCAN mineral mapping system without its complexity and cost constraints. The model exhibited promising performance, and it is expected that the most relevant problems are not related to the methodology itself but to the maps quality and images representativeness, which could be mitigated by, respectively, acquiring higher-resolution maps and increasing the training dataset.

Further investigation could also be conducted by including additional image processing techniques in the workflow. For instance, the QEMSCAN maps could be combined to prior thin section pixels clustering for smoothing and correction of the maps boundaries. Also, instead of transforming the QEMSCAN maps, the thin section images could be downsampled following the maps sampling resolution pattern, guaranteeing a perfect image-map correspondence despite the loss of RGB information.

Overall, it is noticeable that a QEMSCAN-based mineralogical segmentation model has the potential to replace the actual QEMSCAN mapping in the future, with the suitable improvements made. Also, it is worth remembering that a production-level model would be trained with all available data, i.e., including the data designated for validation in the described experiments, reaching its maximum performance.

## 6. Acknowledgments

The authors acknowledge LTrace Geosciences for the technical support and Petrobras for the technical support and funding.

**Code availability section**

Deep Mineralogical Segmentation.

Contact: jean.mello@ltrace.com.br, jeanpvmello@gmail.com or +55 27 99239 8192

Hardware recommendations:

- 8 GB RAM or superior in case of manipulation of high-resolution images;
- GPU for training and evaluation. Not known memory restrictions. Use smaller batch sizes and cache rates for GPUs with less memory.

Program language: Python (tested versions: 3.9.6 and 3.9.12).

Software required: GeoSlicer (for image registration and SOI delimitation).

Program size:

- Data generation, training and evaluation code: ~2 MB;
- GeoSlicer: ~9 GB.

The source code/software are available for downloading at the links:

- Data generation, training and evaluation code: `https://github.com/ltracegeo/deep-mineralogical-segmentation`
- GeoSlicer: `https://github.com/petrobras/GeoSlicer`.

# List of Figures

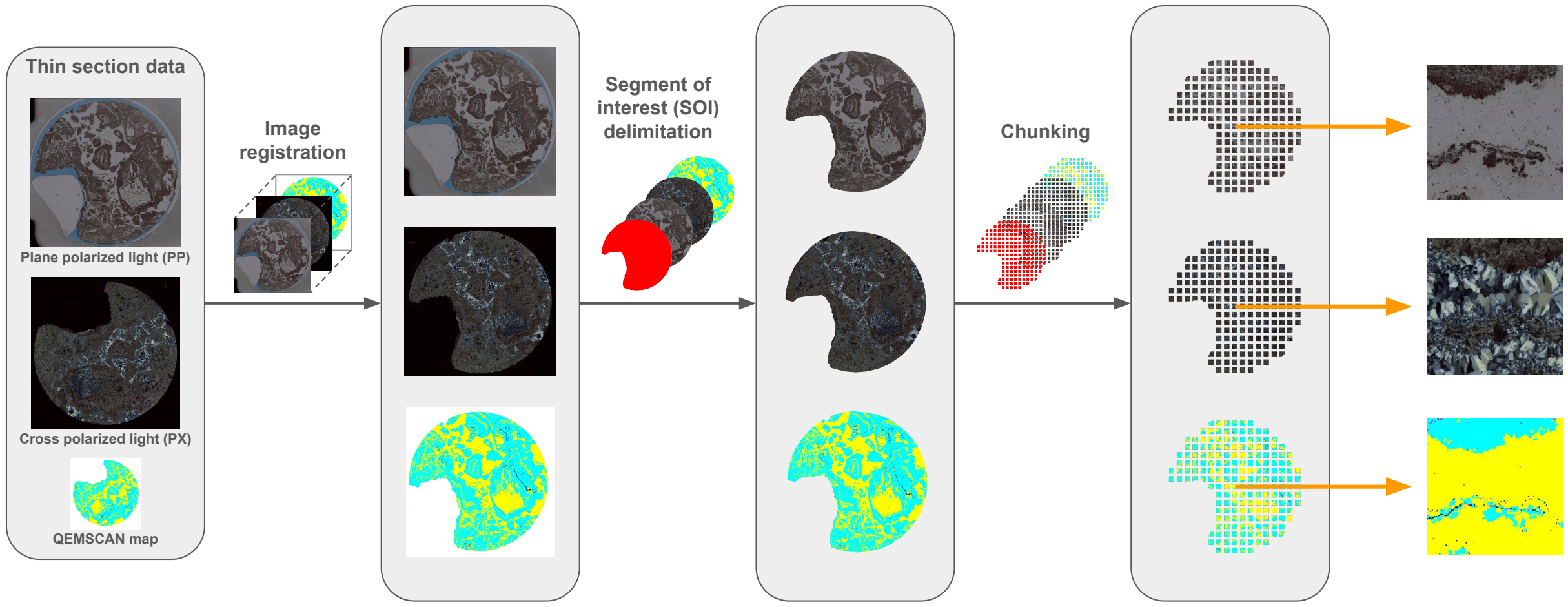

**Figure 1:** Overview of the data aspect and preprocessing. The PP image, XP image and QEMSCAN map from each thin section were subjected to image registration, SOI delimitation and chunking in order to obtain simpler, more numerous and spatially aligned images.

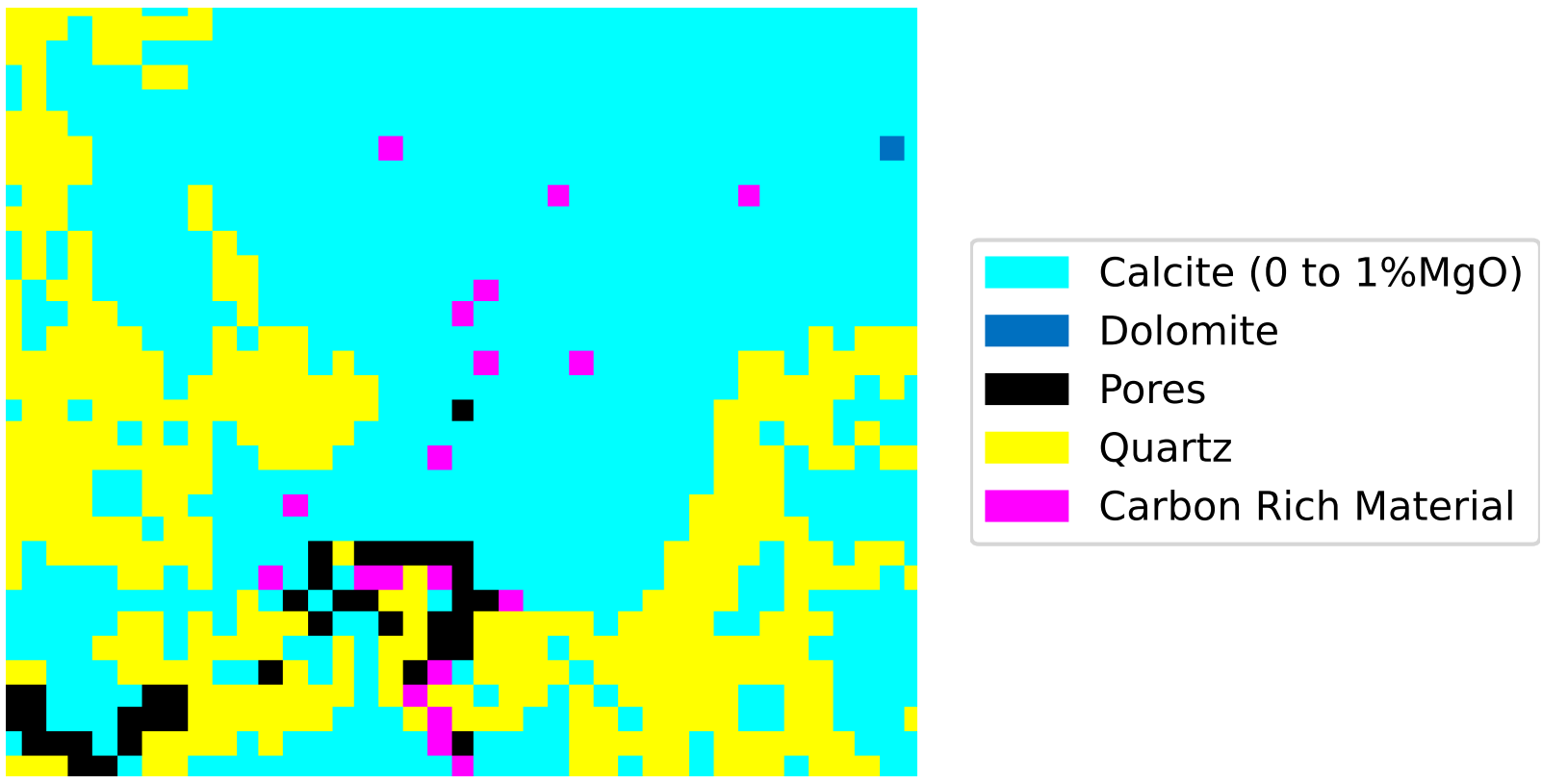

**Figure 2:** Close view of a region from the QEMSCAN map shown in Figure 1. The boundaries between different elements are not as smooth as in the photographs, due to the set SEM sampling resolution.

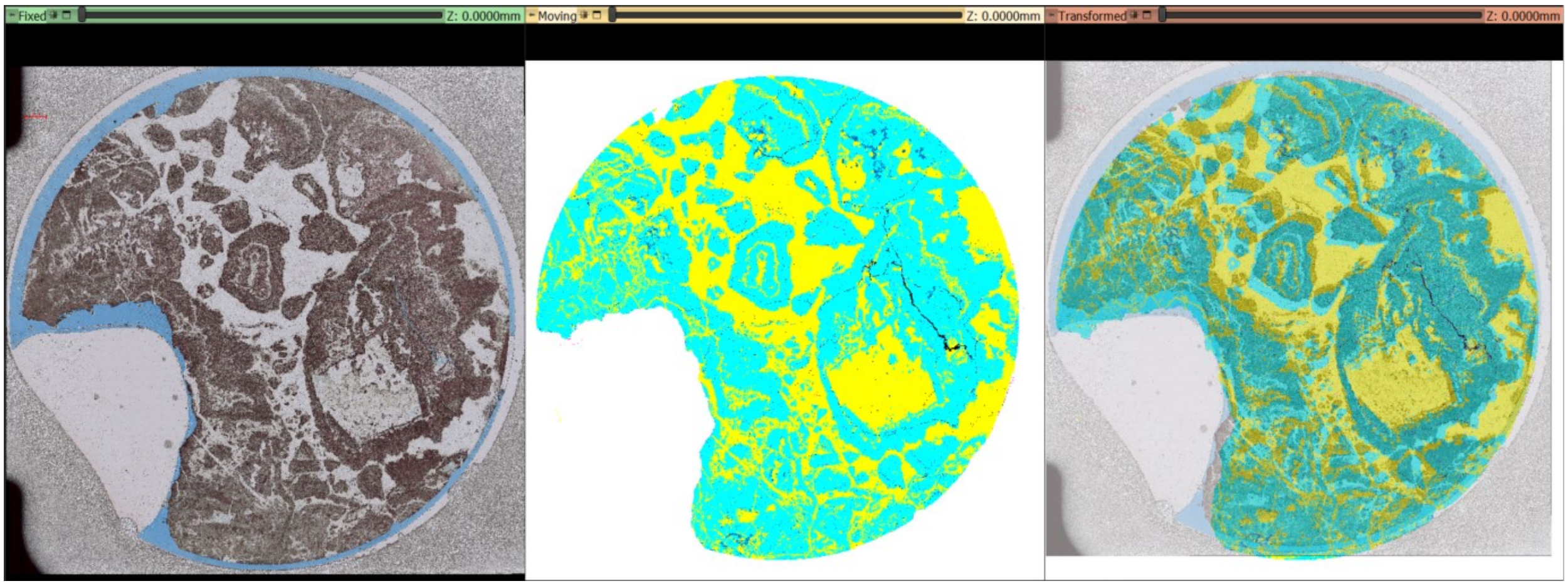

**Figure 3:** PP image (left) and QEMSCAN map (center) loaded and scaled as, respectively, fixed and moving images for registration, along with a visualization of the images overlap before the registration process (right).

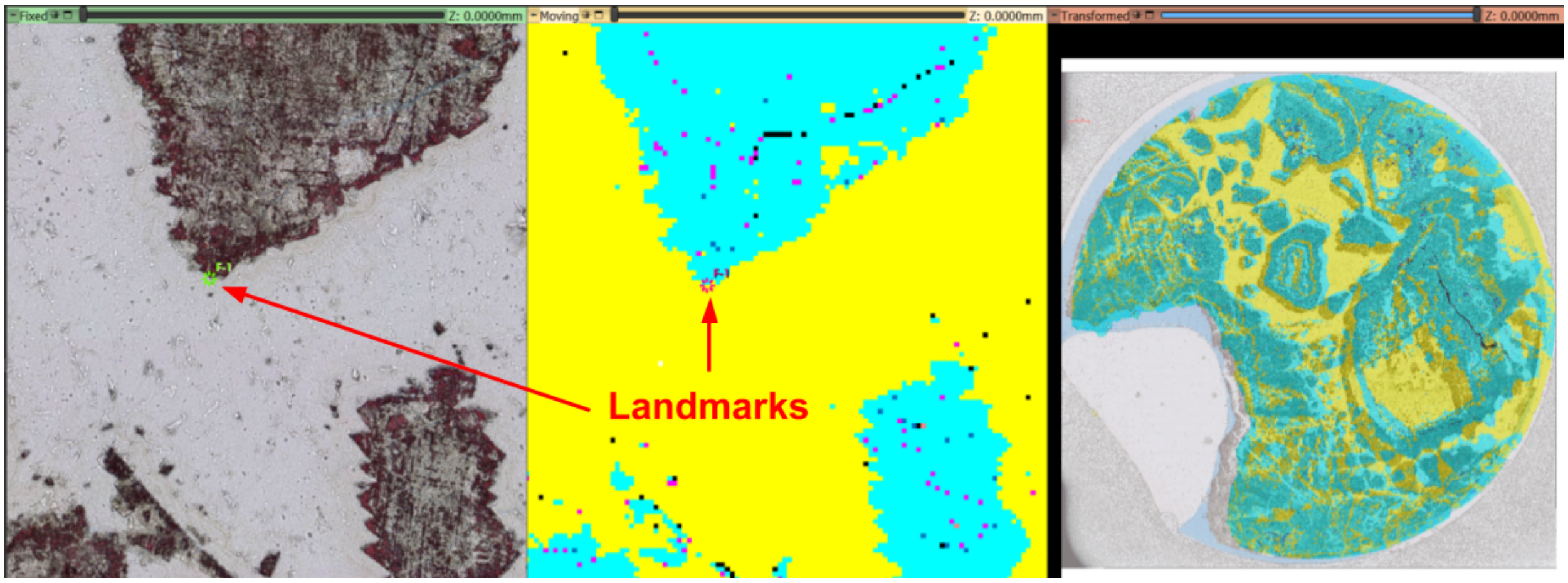

**Figure 4:** First pair of landmarks set in corresponding locations from PP image (left) and QEMSCAN map (center), along with the partial result of registration with this single pair (right).

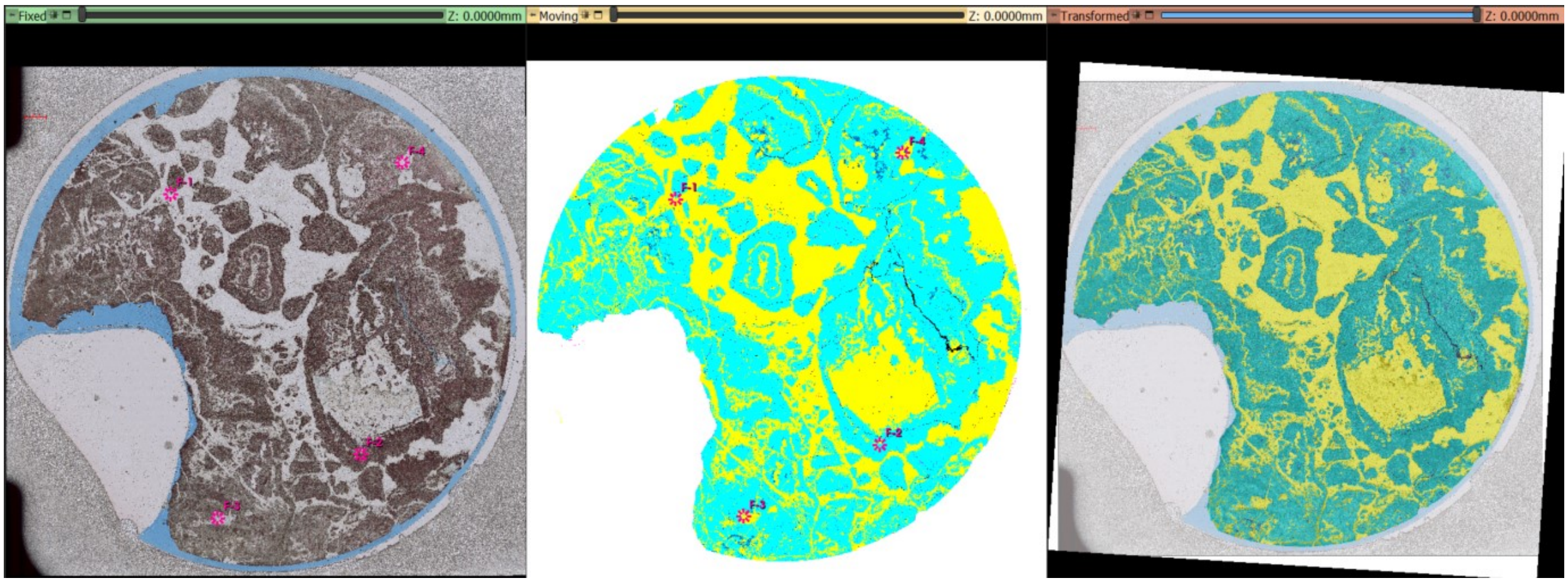

**Figure 5:** PP image (left), QEMSCAN map (center) and the final result of registration with four landmark pairs (right).

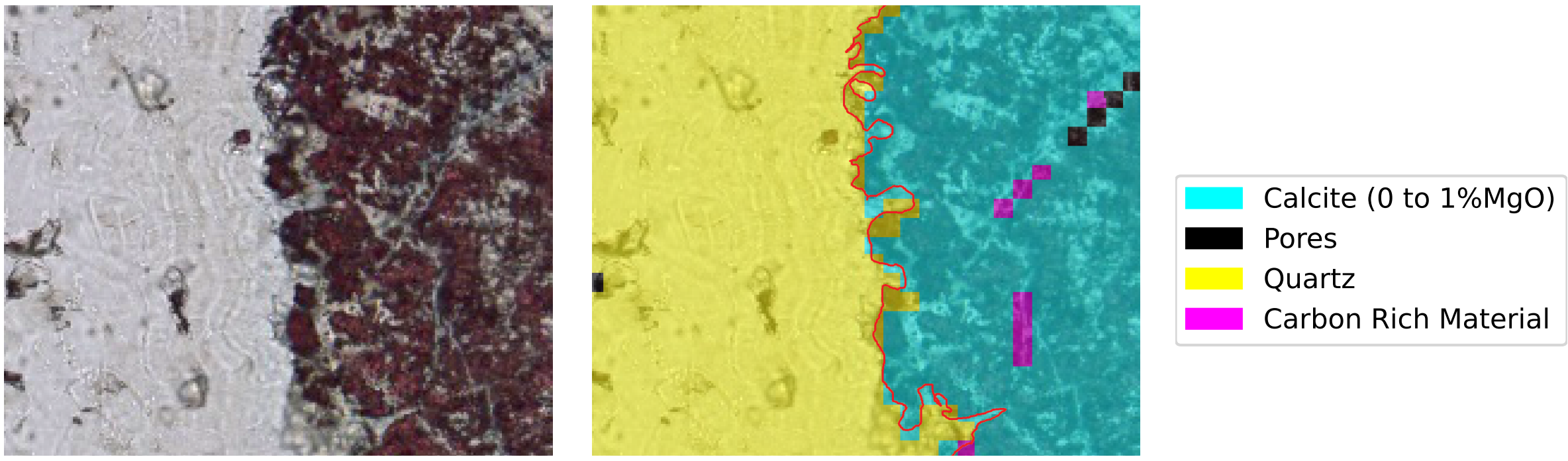

(a) (b)

**Figure 6:** Example of the registration problem: (a) Crop from the PP image; (b) registered map overlapping the crop. The registration problem is evidenced by the difference between the map's quartz-calcite boundary and the estimated real boundary (red scribble). Some small islands of carbon rich material and pores are also visible, presenting rectangular forms which can hardly match their real correspondents in PP.

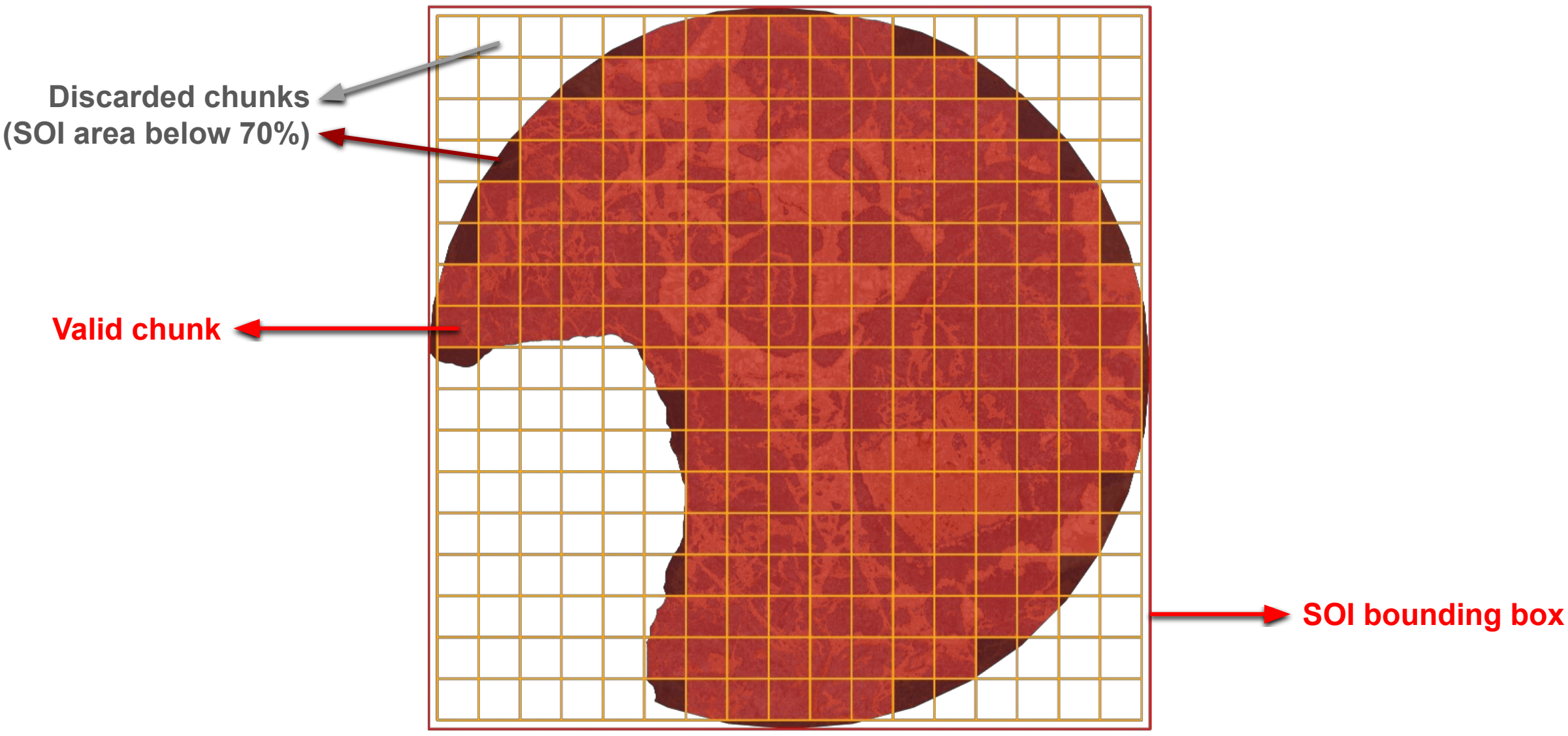

**Figure 7:** The chunking process. The chunks are generated only inside the SOI bounding box's largest center crop that can fit them. Also, all chunks with less than 70% of area covered by SOI are discarded.

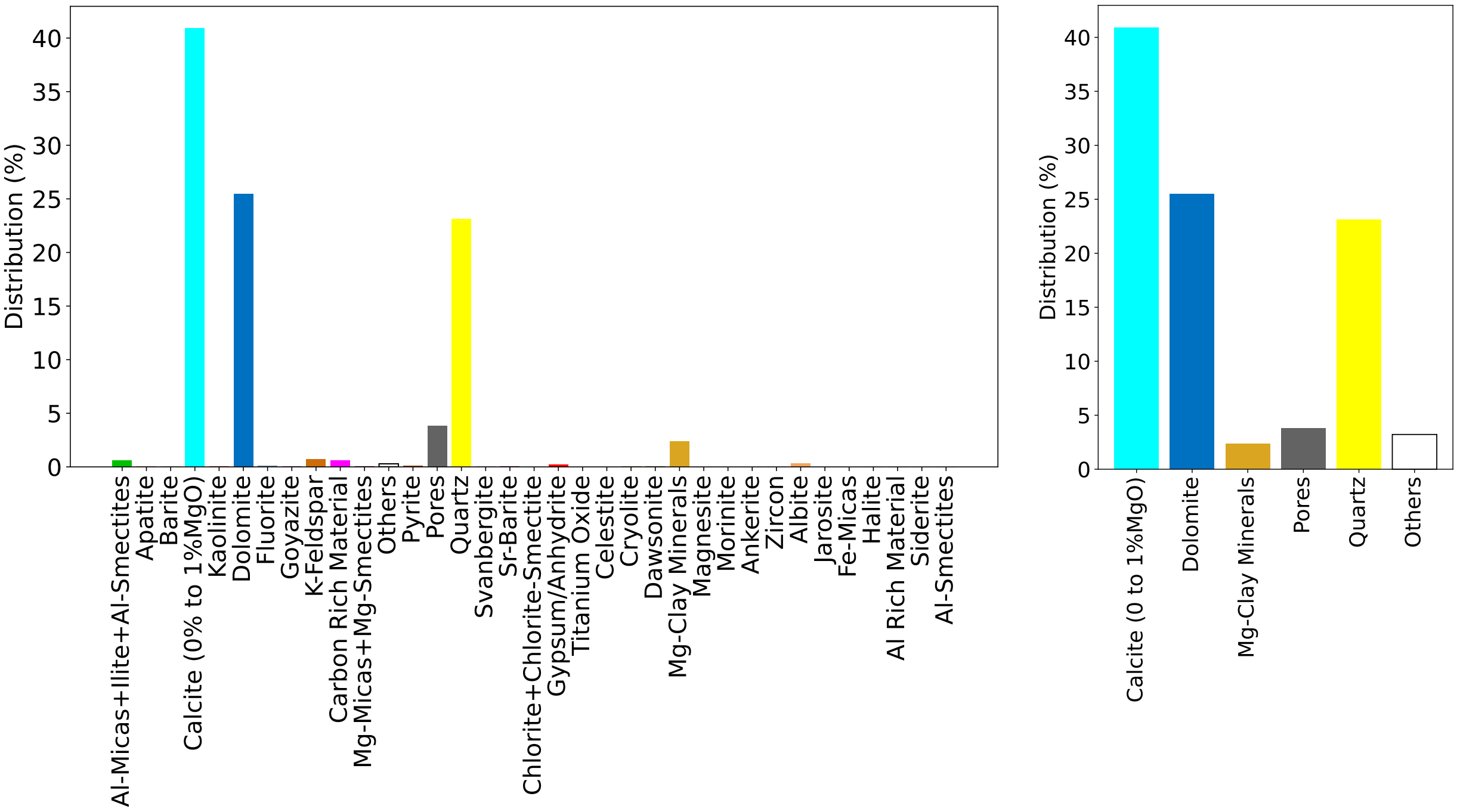

**Figure 8:** At left, the original mineral phases distribution in the final dataset. At right, the distribution after expanding the class "Others" to encompass all but the five most frequent phases.

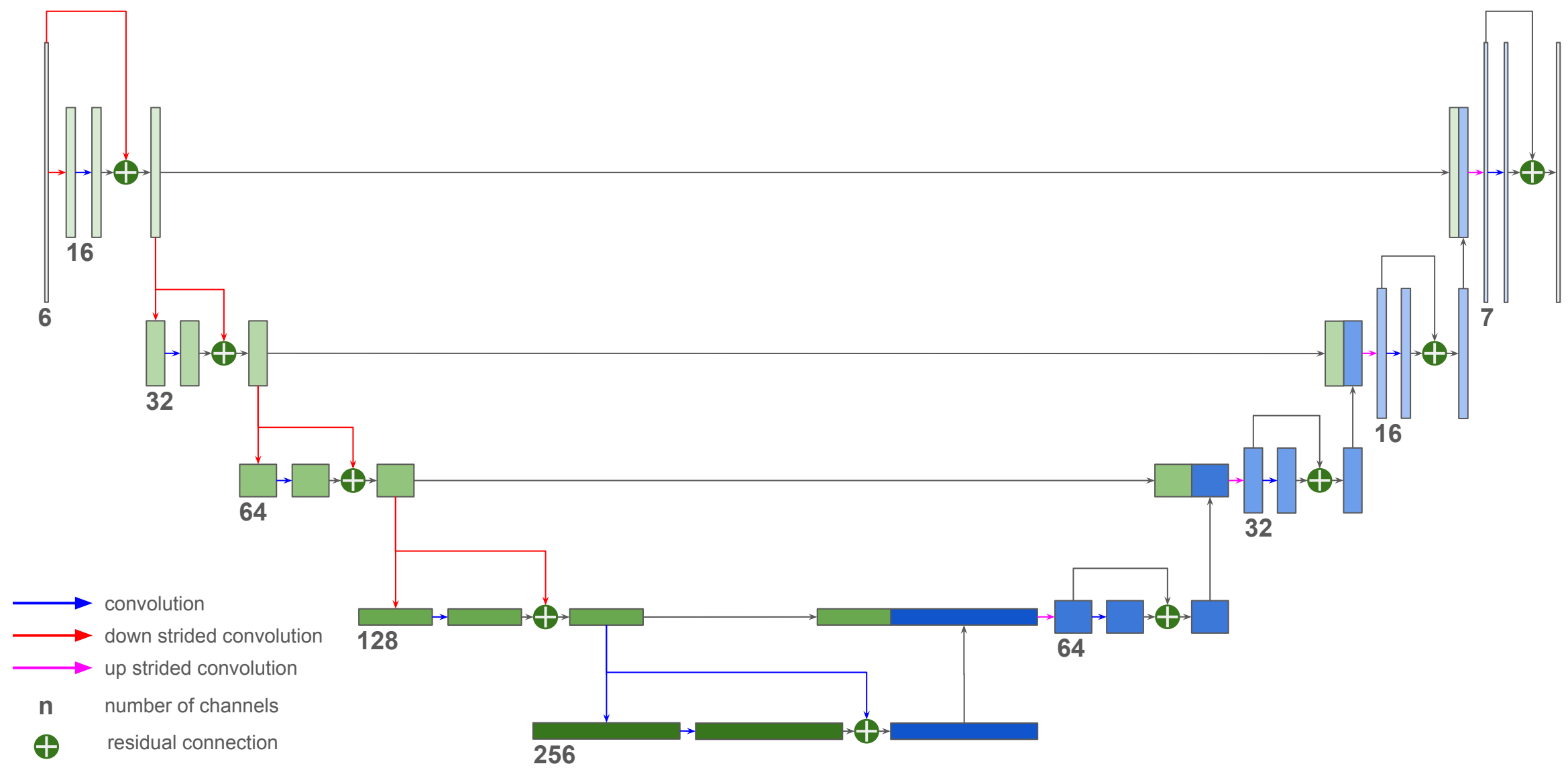

(a)

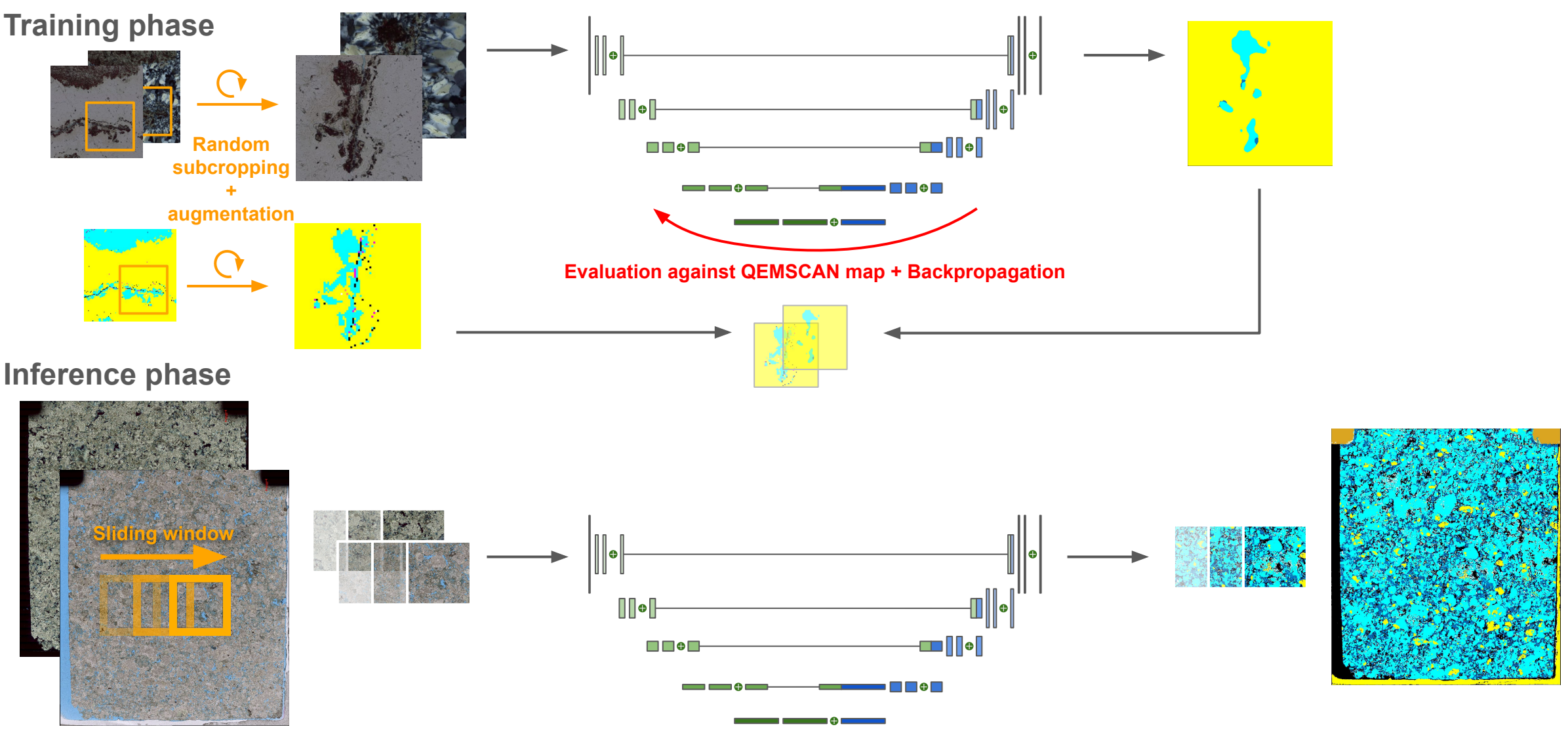

(b)

**Figure 9:** (a) The Residual U-Net architecture used. It uses strided convolutions to encode six-channel input images (PP RGB + XP RGB) into a deep low-resolution feature space and decode it into input-resolution activation maps for each of the six classes plus unlabeled region (outside SOI). Each layer benefits from a residual connection between its first and last feature maps; (b) Examples of training and inference iterations. During training, a subcrop from the input chunk is spatially augmented and the resulting map is compared to the corresponding QEMSCAN map for updating the model's parameters. Inference, in turn, is applied on overlapping windows from the input and the predictions are assembled to build the resulting map for the entire input.

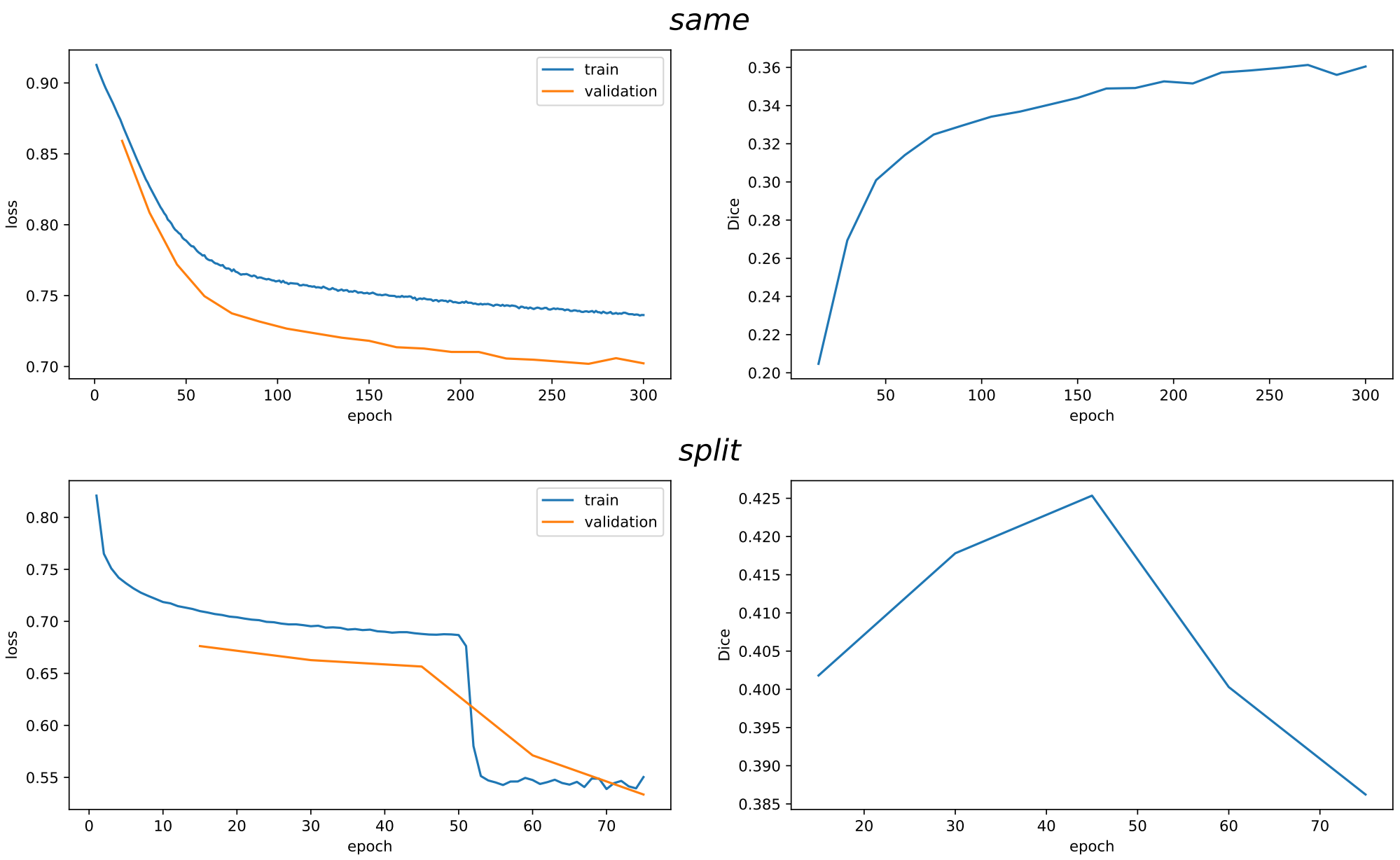

**Figure 10:** Curves of loss in the left and validation Dice scores in the right for the each validation method, *same* in the top and *split* in the bottom.

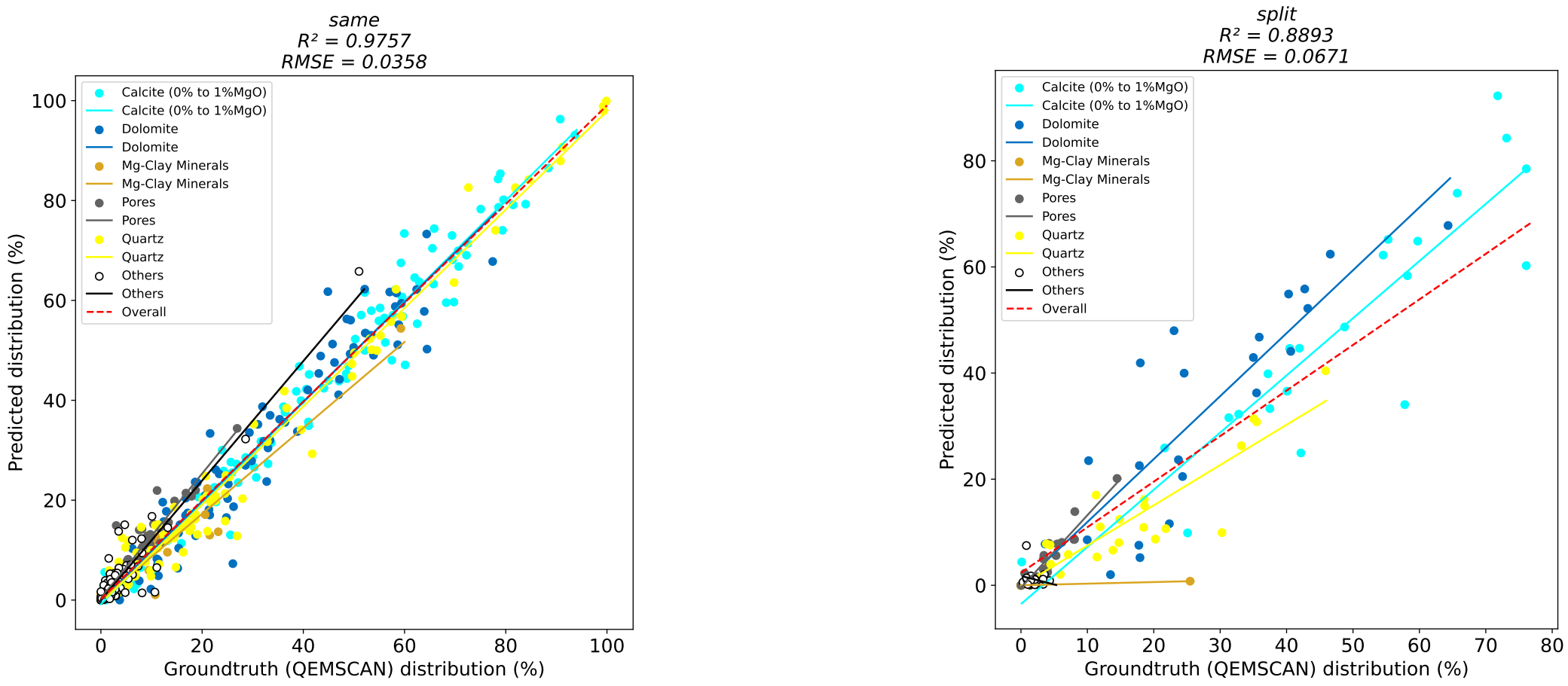

**Figure 11:** Linear correlation between the predicted and groundtruth mineral phases distributions in each validation thin section, both overall and per class, for each validation method.

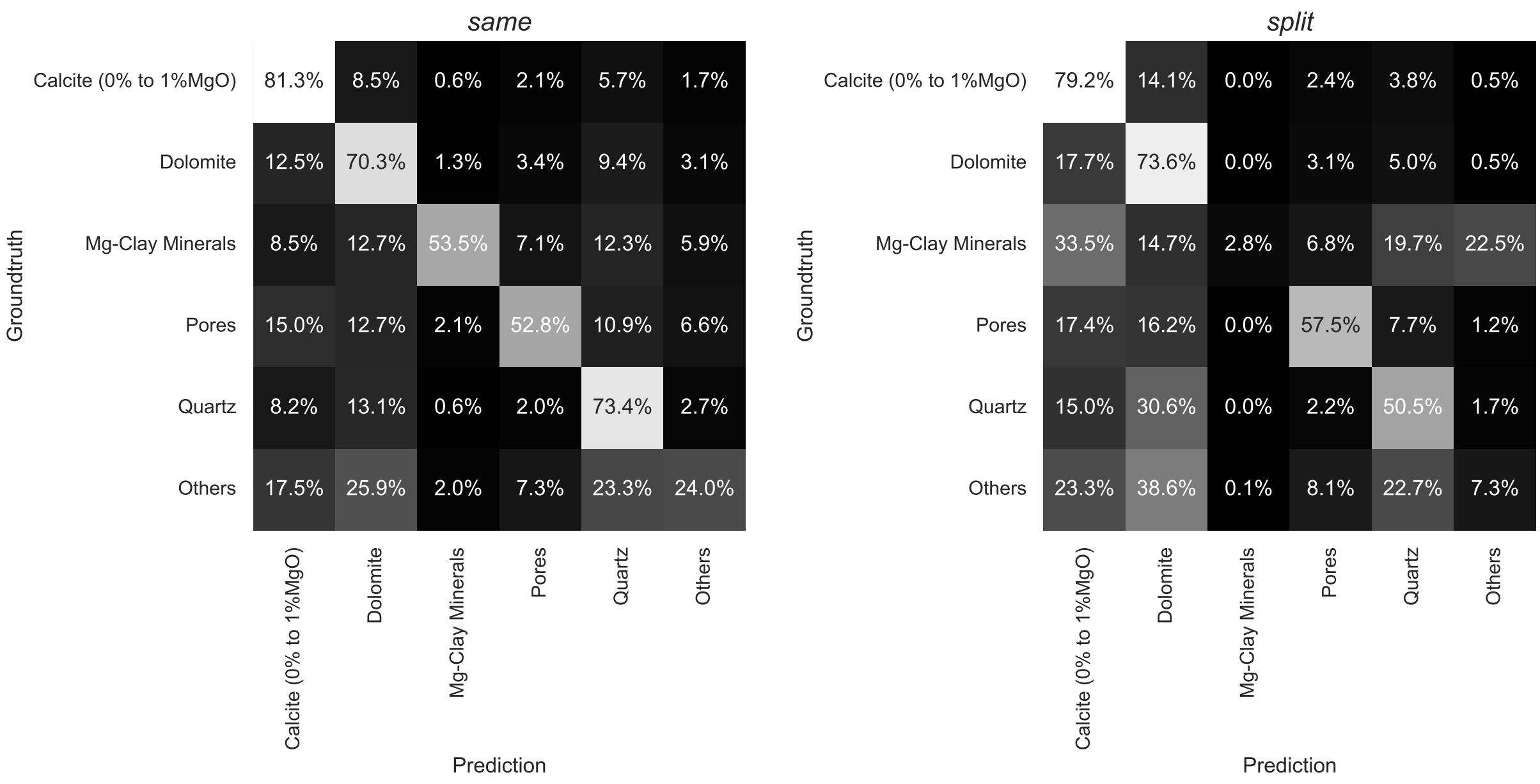

**Figure 12:** Confusion matrix for each validation method.

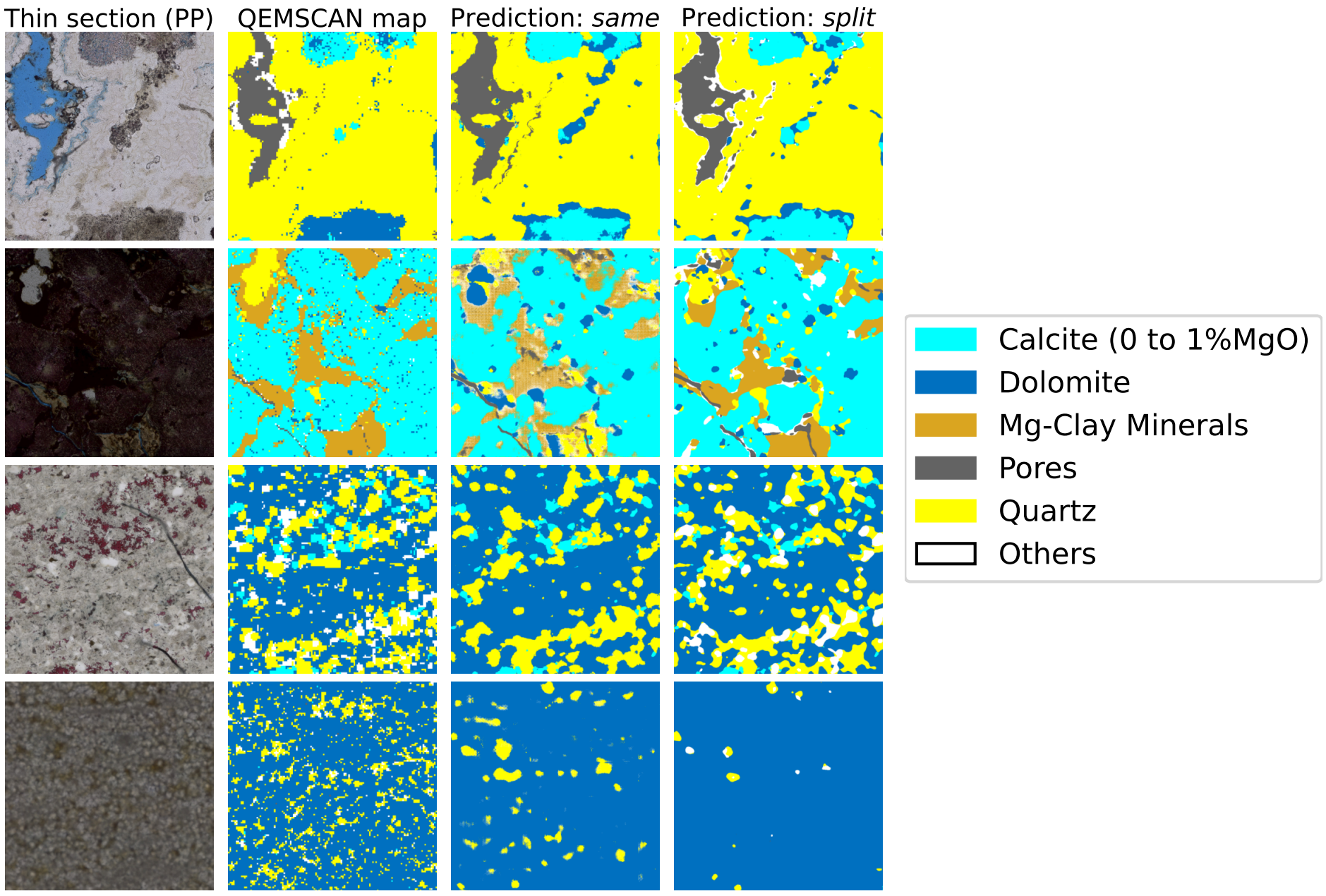

**Figure 13:** Visual prediction examples for each validation method.